\documentclass[fleqn,10pt]{wlscirep}
\usepackage{cite}
\usepackage{graphicx}
\usepackage{textcomp}
\usepackage{amsmath,amssymb,amsfonts}
\usepackage{cleveref}
\usepackage{fancyvrb}
\usepackage{setspace}

\usepackage{booktabs}
\usepackage{multirow}
\usepackage{float}
\usepackage{svg}
\usepackage{mathrsfs}
\usepackage{braket}
\usepackage{graphicx} % Required for inserting images
\usepackage{pgfplots}
\usepackage{subcaption} 
\usepackage{tikz}
\usepackage{multirow}
\usepackage{multicol}
\usepackage{threeparttable} 
\newcommand*\circled[1]{\tikz[baseline=(char.base)]{
            \node[shape=circle,fill,inner sep=0.5pt] (char) {\textcolor{white}{#1}};}}

\usepackage{enumitem}

\usepackage{algorithmic}
\usepackage{graphicx}
\usepackage{textcomp}
\usepackage{xcolor}
\def\BibTeX{{\rm B\kern-.05em{\sc i\kern-.025em b}\kern-.08em
    T\kern-.1667em\lower.7ex\hbox{E}\kern-.125emX}}

\title{Deep Quanvolutional Neural Networks with Enhanced Trainability and Gradient Propagation\\
}

\author[1,2*]{Muhammad Kashif}
\author[1,2]{Muhammad Shafique}
\affil[1]{eBrain Lab, Division of Engineering, New York University Abu Dhabi, PO Box 129188, Abu Dhabi, UAE }
\affil[2]{Center for Quantum and Topological Systems, NYUAD Research Institute, New York University Abu Dhabi, UAE}

\affil[*]{muhammadkashif@nyu.edu}


\begin{abstract}
In this paper, we explore methods to enhance the performance of one of the frequently used variants of Quantum Convolutional Neural Networks, known as Quanvolutional Neural Networks (QuNNs) by introducing trainable quanvolutional layers and addressing the challenges associated with training multi-layered or deep QuNNs. Traditional QuNNs mostly rely on static (non-trainable) quanvolutional layers, limiting their feature extraction capabilities. Our approach enables the training of these layers, significantly improving the scalability and learning potential of QuNNs. However, multi-layered deep QuNNs face difficulties in gradient-based optimization due to limited gradient flow across all the layers of the network.
To overcome this, we propose Residual Quanvolutional Neural Networks (ResQuNNs), which utilize residual learning by adding skip connections between quanvolutional layers. These residual blocks enhance gradient flow throughout the network, facilitating effective training in deep QuNNs, thus \emph{enabling deep learning in QuNNs}. Moreover, we provide empirical evidence on the optimal placement of these residual blocks, demonstrating how strategic configurations improve gradient flow and lead to more efficient training.
Our findings represent a significant advancement in quantum deep learning, opening new possibilities for both theoretical exploration and practical quantum computing applications.

\end{abstract}

\begin{document}

\flushbottom
\maketitle

\section{Introduction}
Quantum Machine Learning (QML) combines quantum computing with machine learning to process and analyze data using quantum algorithms and circuits\cite{Schuld_2014}. By leveraging quantum phenomena such as superposition and entanglement, QML aims to provide computational advantages over classical methods, especially for complex, high-dimensional problems\cite{Biamonte_2017,kashif2025computationaladvantage,kashif:2022_demonstrating}. 
Quantum Convolutional Neural Networks (QCNNs) represent an innovative fusion of quantum computing principles with traditional convolutional neural network (CNN) architectures \cite{wei:2022, zaman2023survey}. These hybrid classical-quantum models utilize parameterized quantum circuits (PQCs) \emph{a.k.a} ansatz, composed of parameterized rotation gates such as $R_x$, $R_y$, or $R_z$, functioning analogously to convolutional and pooling operations in classical CNNs \cite{Cong:2019,Rajesh:2021}. The optimization, however, is performed on classical computers \cite{Hur:2022,Cerezo_2022,Shen:2023,Kashif:2024_alleviating}.
A notable advantage of QCNNs lies in their ability to harness the vast Hilbert space offered by quantum mechanics, which surpasses the capabilities of classical CNNs \cite{Schuld:2019}. This feature allows QCNNs to capture spatial relationships in image data more effectively. Furthermore, the potential future inclusion of higher-level quantum systems, such as qutrits or ququads \cite{sebastian:2023,hu:2023}, promises even more sophisticated image comprehension capabilities.

\begin{figure}[H]
    \centering
    \includegraphics[scale=0.55]{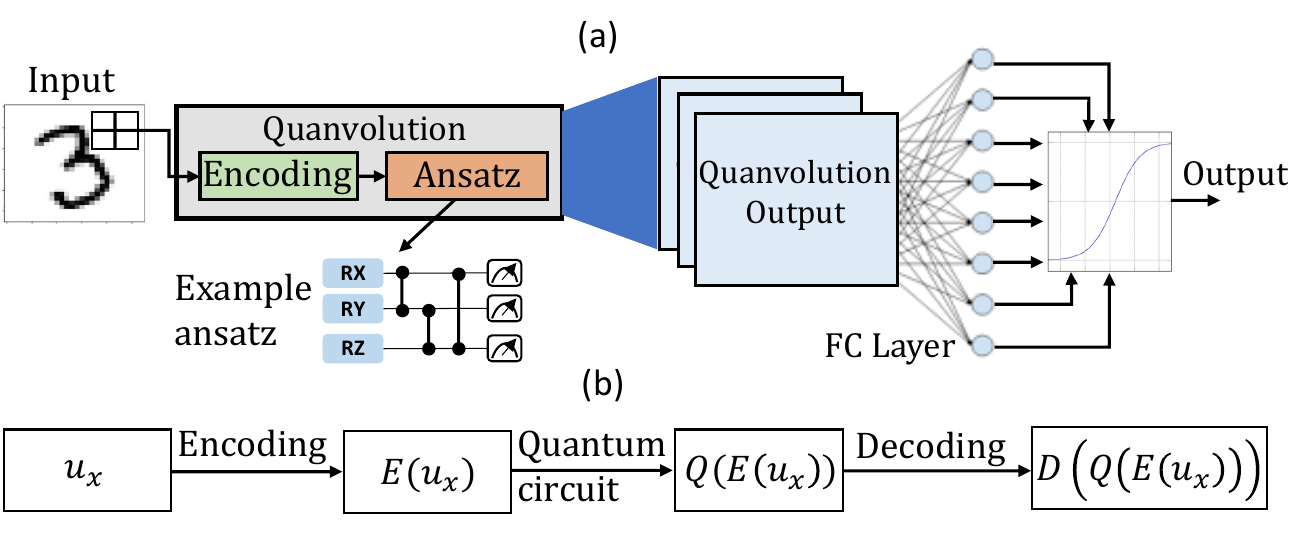}
    % \vspace{-12pt}
    \caption{\footnotesize (a) A typical Illustration of Quanvolutional Neural Network Architecture, (b) Simplified Workflow of Quanvolutional Layer in QuNNs}
    \label{fig:QuNN_architecture}
\end{figure}

There are two primary variants of QCNNs, each leveraging quantum mechanics principles like superposition, entanglement, and interference. The first type, Quantum-Inspired CNNs, proposed in \cite{Cong:2019}, mirrors the structure of classical CNNs but replaces convolutional and pooling layers with deeper PQCs. This design demands higher qubit coherence to fully exploit quantum parallelism and entanglement, posing a challenge for current quantum hardware. Nevertheless, a significant research has been dedicated to explore this very architectural framework for various applications \cite{Hur:2022, Kim:2023}.
The second type, Quanvolutional Neural Networks (QuNNs), first introduced in \cite{henderson:2019}, and visually depicted in Figure \ref{fig:QuNN_architecture}(a), focuses on quantum-enhanced convolutional layers. These networks encode classical image pixels into quantum states, apply unitary transformations via PQCs, and then measure the qubits to create output image channels, which are subsequently processed by classical fully connected layers. 
This approach offers a more feasible implementation with current quantum technology. 
Hence, we also delve into QuNNs with a focus on identifying the challenges faced in training QuNNs and proposing effective solutions to address these issues.

% \vspace{-11pt}

An abstract view of the workflow of quanvolution layer in QuNNs is presented in Figure \ref{fig:QuNN_architecture}(b), which is four step process: 
\textbf{(1)} the original image $U$ is divided into spatially located subsections $u_x$. Each subsection is a $d\times d$ matrix with $d>1$, 
\textbf{(2)} these image subsections are then encoded into a quantum state ($q$), e.g., for an encoding function $E$, the quantum state can be defined as $q_x=E(u_x)$. We encode the input features into the qubit rotation angles using $R_y(\theta)$. 
\textbf{(3)} the encoded state $q_x$ then undergoes a unitary transformation via a quantum circuit \emph{a.k.a} ansatz $Q$ which can be described as; $f_x=Q(q_x)=Q(E(u_x))$,
\textbf{(4)} decoding is then performed via measurements which can be described as, $D(f_x)=D(Q(E(u_x)))$. 
By repeating these steps for different subsections, the full input image is traversed, producing an outcome organized as a multi-channel image.
The output of this quanvolution is usually processed by a classical neural network to produce the final output.

The QuNN architecture depicted in Figure \ref{fig:QuNN_architecture}, widely adopted in state-of-the-art research for various applications such as 2D and 3D radiological image classification \cite{Matic:2022}, high energy physics data analysis \cite{Chen:2022}, and object detection and classification \cite{Meedinti:2023, Baek:2022}. 
In these contexts, the quantum circuit in QuNNs primarily functions as a feature extractor and is not trained during the quanvolution operation. 
Consequently, the learning process is primarily driven by the subsequent classical network. 
Although there are recent studies that employ trainable quantum circuits in quanvolutional layers, to the best of our knowledge, no existing literature addresses multi-layered QuNNs. Table \ref{tab:Table1a} some summarizes recent works utilizing QuNNs with both trainable and non-trainable quanvolutional layers. To achieve full scalability in QuNNs, it is crucial to ensure the trainability of underlying quanvolutional layers in multi-layered architectures.

% While this architecture is standard in many state-of-the-art works, and is widely being used for various applications such as 2D and 3D radiological image classification \cite{Matic:2022}, analysing the high energy physics data \cite{Chen:2022}, for object detection and classification \cite{meedinti:2023, baek:2022}, the quantum circuit used in the QuNNs is primarily used for feature extraction and is not trained during the quanvolution operation. Consequently, the learning process is primarily driven by the subsequent classical network.
% Although, there have some recent works where a trainable quantum circuit is used in this step but to the best of our knowledge no work in the literature works with multi-layered QuNNs. A summary of some recent works using QuNNs with both trainable and untrainable quanvoluional layer is presented in Table \ref{tab:Table1a})
% For QuNNs to be fully scalable, ensuring the trainability of underlying quanvolutional layers in multi-layered QuNNs is crucial.

\begin{table}[htbp]
    \centering
    \caption{Recent State-of-the-art QuNNs employing trainable and untrainable quanvolution layers }
    \begin{tabular}{p{2.5cm}|p{5cm}|p{5cm}}
    \hline
        \hfil Ref & \hfil Trainable Quanvolutional Layers  & \hfil Multiple Quanvolutional Layers \\
        \hline
        \hfil \cite{matondo:2024} &\hfil $\times$  & \hfil $\times $\\
        \hline
        \hfil \cite{ahalya:2023} & \hfil $\times$  & \hfil $\times$ \\
        \hline
        \hfil \cite{sebastianelli:2024} & \hfil $\times$  & \hfil $\times$ \\
        \hline
        \hfil \cite{maouaki:2024} & \hfil $\times$  & \hfil $\times$ \\
        \hline
        \hfil \cite{maouaki:2024advqunn} & \hfil $\times$  & \hfil $\times$ \\
        \hline
        \hfil \cite{vu:2024} & \hfil $\times$  & \hfil $\times$ \\
        \hline
         \hfil\cite{bhatia:2023} & \hfil $\checkmark$  & \hfil $\times$ \\
        \hline
         \hfil\cite{mattern:2021} & \hfil $\checkmark$  & \hfil $\times$ \\
         \hline
         \hfil \textbf{Ours} & \hfil $\checkmark$  & \hfil $\checkmark$ \\
        \hline
    \end{tabular}
    
    \label{tab:Table1a}
\end{table}

%%%%%%%%%%%%%%%%%%%%%%%%%%%%%%%%%%%%%%%%%%%%%%%%%%%%%%%%%%%%%%%%%%%%%%%%%%%%%%%%%%%%%%%%%%%%%%%%%%%%%%%%%%%%%%%%%%%%%%%%%%%%%%%%%%%%%%%%%%%%%%%%%%%%%%%%%%%%%%%%%%%%%%%%%%%%%%%%%%%%%%%%%%%%%%%%%%%%%%%%%%%%%%%%%%%%%%%%%%%%%%%%%%%%%%%%%%%%%%%%%%%%%%%%%%%%%%%%%%%%%%%%%%%%%%%%%%%%%%%%%%%%%%%%%%%%%%%%%%%%%%%%%%%%%%%%%%%%%%%%%%%%%%%%%%%%%%%%%%%%%%%%%%%%%%%%%%%%%%%%%%%%%%%%%%%%%%%%%%%%%%%%%%%%%%%%%%%%%%%%%%%%%%%%%%%%%%%%%%%%%%%%%
% \vspace{-14pt}
\subsection{Motivational Case Study} \label{sec:motiv_anaysis}

\vspace{-0.1pt}

\begin{figure*}[htbp]
    \centering
    % \hspace{-2pt}
    % height=1in, width=5in
    \includegraphics[scale=0.65]{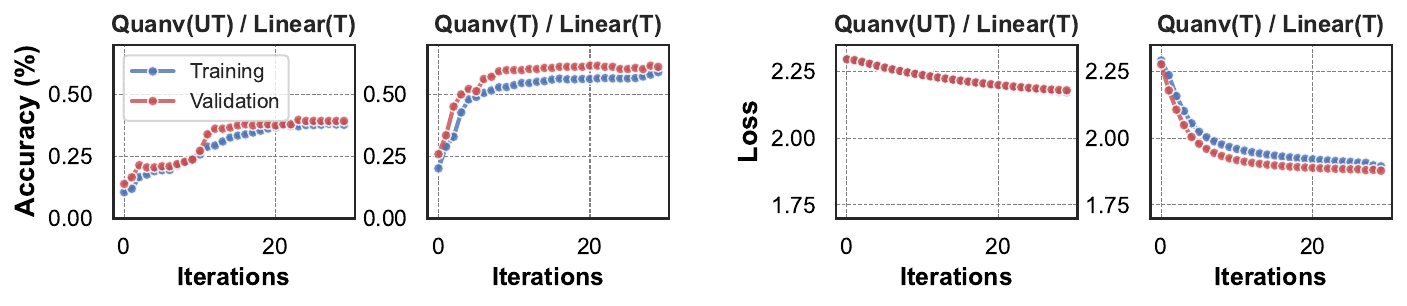}
    % \vspace{-13pt}
    \caption{\footnotesize Motivational analysis showing the impact \emph{Trainable} quanvolutional layers. When the quantunvolutional layer is untrainable (UT) and only the classical linear layer is trained (T), the QuNN yields suboptimal performance. When the quanvolutional layer is made trainable, it significantly enhanced the training performance. Quanv = Quanvolutional layer, Linear = fully connected layer at the end.}
    \label{fig:motiv_analysis}
\end{figure*}

In Figure \ref{fig:motiv_analysis}, we assess the impact of trainable versus untrainable quanvolutional layers in QuNNs.
To this end, we train QuNNs with both trainable (allowing weight updates during backpropagation) and untrainable (frozen weights, no updates) quanvolutional layers. This training is performed on $1000$ randomly chosen images from MNIST dataset over $30$ iterations, using an $80\%-20\%$ training-validation split. 
The Adam optimizer with a $0.01$ learning rate was employed. 
We utilized a $2\times 2$ kernel resulting in $4$-qubit quanvolutional layers, with each layer comprising of $4$ parameterized $R_y$ gates and $3$ $CNOT$ gates for (nearest neighbor) qubit entanglement. The overall depth of quantum circuit in the quanvolutional layer is set to $2$, accounting to a total of $8$ single and $6$ two-qubit gates.

%figure moved above text for better placement in PDF

Results showed suboptimal performance when quanvolutional layers were untrainable and only the final fully connected (FC) layer was trained. Conversely, allowing weight updates in the quanvolutional layers significantly improved learning, enhancing model accuracy by approximately $36\%$. 
% This finding underscores the importance of trainable quanvolutional layers in enhancing QuNNs' overall effectiveness.
\textit{This empirical evidence robustly substantiates the hypothesis that facilitating the trainability of quanvolutional layers significantly augments the model's overall learning efficacy.}
% as we notice an improvement in performance by approximately $36\%$ in terms of achieved accuracy.
%
% \vspace{-5pt}
\begin{figure}[htbp]
    \centering
    \includegraphics[scale=0.5]{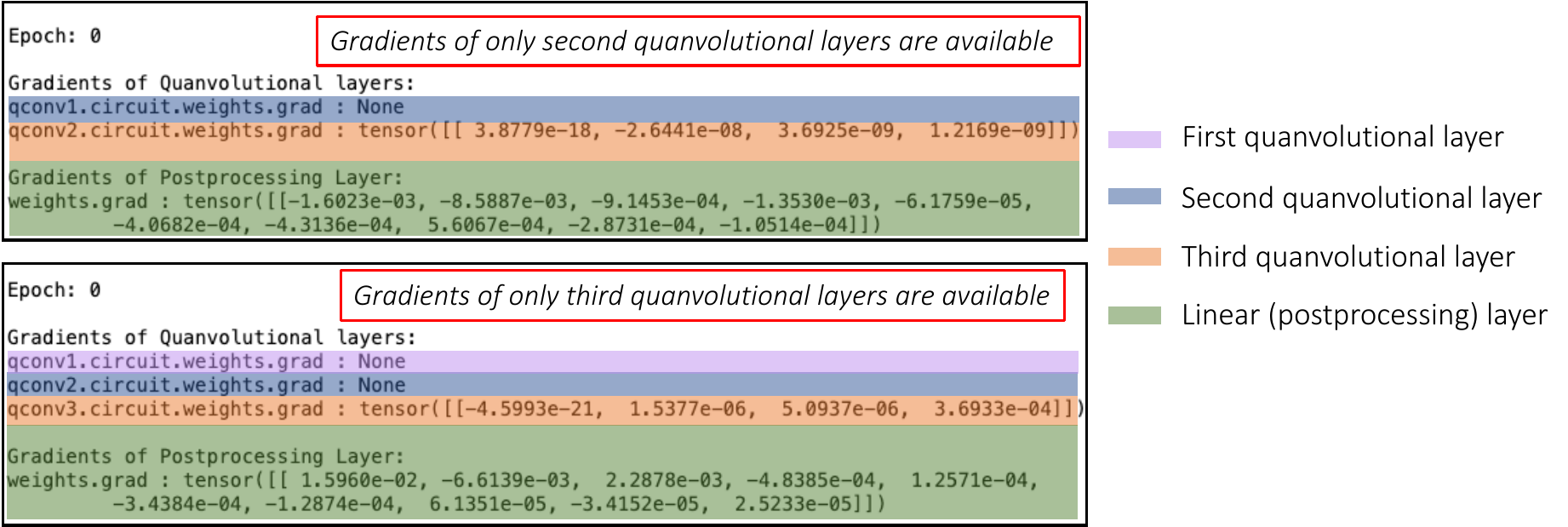}
    % \vspace{-5pt}
    \caption{\footnotesize Gradients Propagation in multi-layered QuNNs with Two (upper pannel) and Three (lower panel) Quanvolutional Layers.}
    \label{fig:motiv_analysis_Grads}
\end{figure}

% \vspace{-7pt}

While a single trainable quanvolutional layer enhances QuNNs' training efficacy, as depicted in Figure \ref{fig:motiv_analysis}, scalability of QuNNs requires multiple trainable layers. 
% Also, the rationale for using multiple layers is to avoid the problem of vanishing gradients a.k.a \emph{Barren Plateaus} (BP), common in single, more expressive PQCs with a higher qubit count and sufficient depth\cite{McClean:2018,Kashif:2023, kashif2023alleviating}. Multiple layers, with progressively smaller quantum circuits, can potentially mitigate the BPs\cite{kashif2023resqnets}.
However, challenges emerge when multiple quanvolutional layers are trained concurrently, as gradients are exclusively accessible only in the last quanvolutional layer, thus limiting the optimization process only to this layer. The gradient accessibility in multi-layered QuNNs is hindered because each quantum layer performs qubit measurement, thus collapsing the quantum state and breaking the chain of differentiability.
Figure \ref{fig:motiv_analysis_Grads} demonstrates this by showing gradients for two and three trainable quanvolutional layers.
We observe that the gradients of only the final quanvolutional layer are accessible, while those of the preceding layers are effectively \emph{None}. 
The restricted gradient flow means that in multi-layered QuNNs, only the last quanvolutional layer contributes in the learning process, leaving earlier layers remain uninvolved in optimization. This presents a significant challenge in effectively training multi-layered QuNNs, thereby necessitating innovative approaches to address this challenge for scalable and effective QuNN architectures.
% \vspace{-10pt}

% \begin{figure}[htbp]
%     % \centering
%     \hspace{-8cm}
%     \begin{subfigure}
%         \centering
%         \includegraphics[height=0.65in, width=3.2in]{./Figures/2QCL_NR_grads.pdf}
%         \label{fig:2QCL_NR_grads}
%     \end{subfigure}%
    
%     \hspace{-8cm}
%     \begin{subfigure} 
%         \centering
%         \includegraphics[height=0.65in, width=3.2in]{./Figures/3QCL_NR_grads.pdf}
%         \label{fig:3QCL_NR_grads}
%     \end{subfigure}
%     \vspace{-0.7cm}
%     \caption{Gradients Propagation for Two (upper pannel) and Three Quanvolutional Layers (lower panel)}
%     \label{fig:motiv_analysis_Grads}
% \end{figure}

% \vspace{-0.35cm}

% \vspace{-10pt}

% \vspace{-12pt}

% \vspace{-7pt}

%%%%%%%%%%%%%%%%%%%%%%%%%%%%%%%%%%%%%%%%%%%%%%%%%%%%%%%%%%%%%%%%%%%%%%%%%%%%%%%%%%%

\subsection{Our Contributions}
\begin{figure}[htbp]
    \centering
    % \hspace{-15pt}
    \includegraphics[scale=0.5]{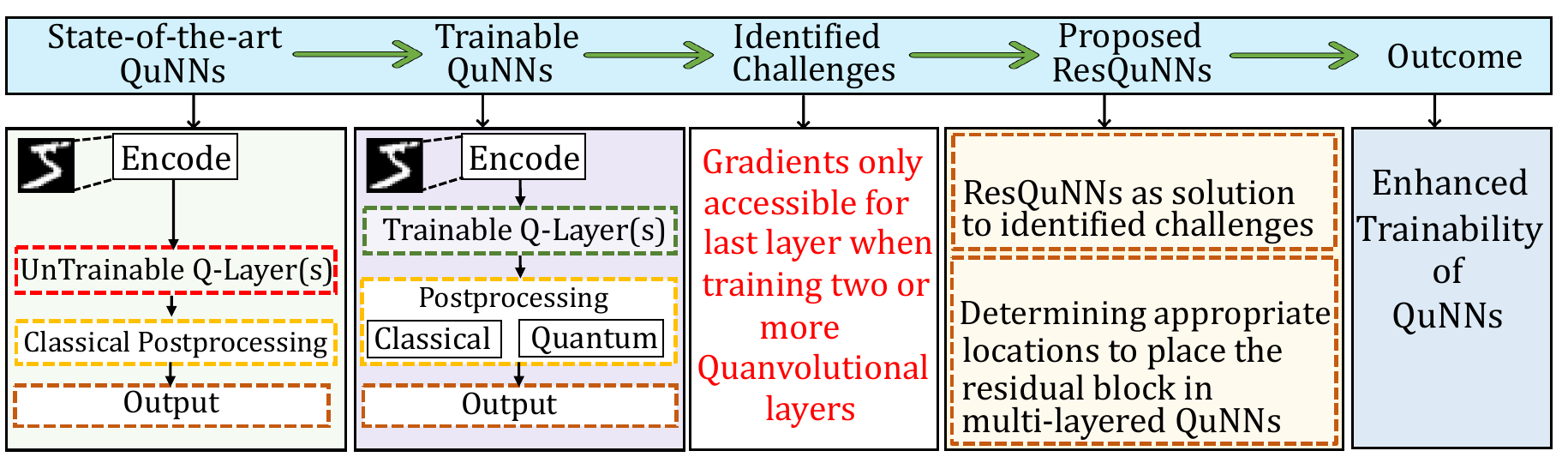}
    
    \caption{\footnotesize Overview of our contributions }
    \label{fig:contributions}
\end{figure}

Our primary contributions encompass the introduction of trainable quanvolutional layers, identification of challenges with multiple quanvolutional layers, the development of Residual Quanvolutional Neural Networks (ResQuNNs) to address these challenges, and empirical findings on the strategic placement of residual blocks to further enhance training performance.
An overview of our contributions is shown in Figure \ref{fig:contributions}.

\begin{itemize}%[leftmargin=-.075in]
    \item \textbf{\textit{Trainable Quanvolutional Layers.}} We first make the quanvolutional layers trainable in QuNNs, addressing the challenge of enhancing their adaptability within QuNNs and improving their training performance.

    \item \textbf{\textit{Challenges with Multiple Quanvolutional Layers.}} We identify a key challenge in optimizing multiple quanvolutional layers, highlighting the complexities associated with accessing gradients across these layers. In QuNN architectures with multiple trainable quanvolutional layers, the gradients do not flow through all the layers of the network, making deeper QuNNs unsuitable for gradient-based optimization techniques.

    \item \textbf{\textit{Residual Quantum Convolutional Neural Networks (ResQuNN).}} To overcome the gradient accessibility issue, this paper introduces Residual Quantum Convolutional Neural Networks (ResQuNN), drawing inspiration from classical residual neural networks.

    \item \textbf{\textit{Residual Blocks for Gradient Accessibility.}} In ResQuNNs, the residual blocks between quanvolutional layers are incorporated to facilitate comprehensive gradient access throughout the network, thereby enabling the deep learning in QuNNs and eventually improving training performance of QuNNs. These residual blocks combine the output of existing quanvolutional layers with their own or previous layer's input before forwarding it to the subsequent quanvolutional layer.

    \item \textbf{\textit{Strategic Placement of Residual Blocks.}} We conducted extensive experiments to determine appropriate locations for inserting residual blocks in ResQuNNs which ensure comprehensive gradient throughout the network. 

    \item \textbf{\textit{Classical and Quantum Postprocessing.}} State-of-the-art QuNNs utilize classical networks to postprocess the output of quanvolutional layers. However, to demonstrate the effectiveness of the residual approach in multi-layered QuNNs (with no influence on the overall learning from classical layers), other than classical postprocessing layer, we employ the quantum circuit itself for postprocessing the results of the quanvolutional layers.
    
\end{itemize}

%%%%%%%%%%%%%%%%%%%%%%%%%%%%%%%%%%%%%%%%%%%%%%%%%%%%%%%%%%%%%%%%%%%%%%%%%%%%%%%%%%%%%%%%%%%%%%%%%%%%%%%%%%%%%%%%%%%%%%%%%%%%%%%%%%%%%%%%%%%%%%%%%%%%%%%%%%%%%%%%%%%%%%%%%%%%%%%%%%%%%%%%%%%%%%%%%%%%%%%%%%%%%%%%%%%%%%%%%%%%%%%%%
\section{Related Work}

Recently, some studies have explored the application of residual connections in quantum neural networks (QNNs). In \cite{kashif2023resqnets}, residual connections are incorporated into a simple feedforward QNN architecture to address the problem of barren plateaus (BPs). BPs are characterized by the exponential vanishing of gradients in parameterized quantum circuits, which are essential components of QNNs, as the number of qubits increases. By partitioning the conventional QNN architectures into multiple nodes, and then by adding residual connections between these nodes, \cite{kashif2023resqnets} successfully mitigates the BP issue.
Our work diverges from \cite{kashif2023resqnets} in two significant ways: (1) We employ residual approach in a different architecture, specifically QuNNs, and (2) we address a distinct underlying problem, i.e., restricted gradient flow in QuNNs, which occurs regardless of the number of qubits unlike BPs which occurs with increasing number of qubits. To overcome the partial gradient flow in QuNNs, we apply residual connections.

In another recent work \cite{wen:2024}, the concept of residual connections is used in QNNs. The proposed quantum residual neural networks (QResNets) leverage auxiliary qubits in data-encoding and trainable blocks to extend frequency generation forms and improve the flexibility of adjusting Fourier coefficients, leading to enhanced spectral richness and expressivity of parameterized quantum circuits at the cost of greater qubit utilization in the form of auxiliary qubits, which is not suitable for NISQ devices. In contrast, our work focuses on QuNNs, addressing the critical challenges of restricted gradient flow in deep, multi-layered QuNNs without using any additional qubits. We introduce trainable quanvolutional layers, significantly increasing the scalability and learning potential of QuNNs. To resolve complexities in gradient-based optimization for QuNNs with multiple layers, we propose ResQuNNs that utilize residual learning and skip connections to facilitate gradient flow across multiple trainable quanvolutional layers. 
While QResNets enhance frequency generation and flexibility in quantum circuits, our work specifically targets the scalability and learning potential of QuNNs, providing practical solutions for efficient training in deep quantum neural networks. 
% This marks a substantial step forward in quantum deep learning, offering new avenues for theoretical development and practical quantum computing applications

%%%%%%%%%%%%%%%%%%%%%%%%%%%%%%%%%%%%%%%%%%%%%%%%%%%%%%%%%%%%%%%%%%%%%%%%%%%%%%%%%%%%%%%%%%%%%%%%%%%%%%%%%%%%%%%%%%%%%%%%%%%%%%%%%%%%%%%%%%%%%%%%%%%%%%%%%%%%%%%%%%%%%%%%%%%%%%%%%%%%%%%%%%%%%%%%%%%%%%%%%%%%%%%%%%%%%%%%%%%%%%%%%
\section{ResQuNN Methodology}

\begin{figure*}[htbp]
    % \centering
    \hspace{-28pt}
    \includegraphics[scale=0.43]{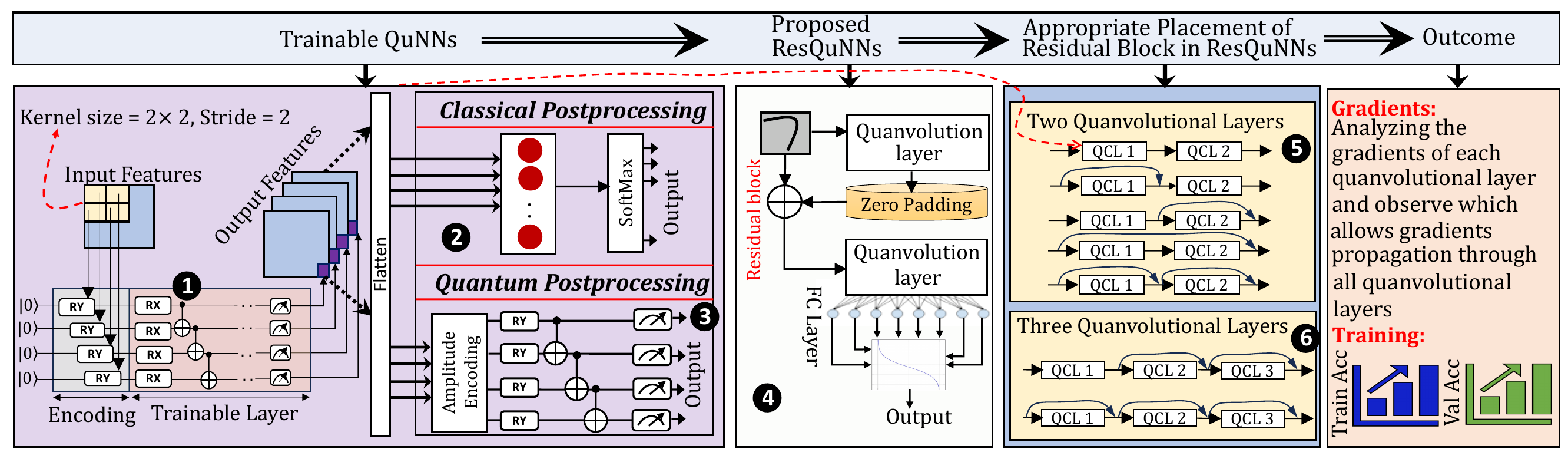}
    % \vspace{-12pt}
    \caption{\footnotesize Detailed Methodology. A comprehensive overview of steps involved in the analysis of trainable QuNNs and proposed ResQuNNs. The results of quanvolutional layers are postprocess using both classical layer and quantum circuit. \emph{Zero Padding} is performed to match the dimensions of input and output before passing it to residual block. The gradients accesibility through all the layers and trainaing and validation accuracy are used as eveluation metrics. QCL = Quanvolutional Layer}
    \label{fig:methodology}
\end{figure*}

% The core objectives revolve around enhancing the adaptability of QuNNs through the introduction of trainable quanvolutional layers, addressing the challenges encountered when optimizing multiple quanvolutional layers, and ultimately proposing a novel architecture known as Residual Quantum Convolutional Neural Networks (ResQuNNs) to overcome the challenges. 
% Additionally, we investigate the strategic placement of residual blocks to maximize training performance within networks containing multiple quanvolutional layers. 
The detailed overview of ResQuNN's methodology is presented in Figure \ref{fig:methodology}. Below we discuss different steps of our methodology in detail.

%%%%%%%%%%%%%%%%%%%%%%%%%%%%%%%%%%%%%%%%%%%%%%%%%%%%%%%%%%%%%%%%%%%%%%%%%%%%%%%%%%%%%%%%%%%%%%%%%%%%%%%%%%%%%%%%%%%%%%%%%%%%%%%%%%%%%%%%%%%%%%%%%%%%%%%%%%%%%%%%%%%%%%%%%%%%%%%%%%%%%%%%%%%%%%%%%%%%%%%%%%%%%%%%%%%%%%%%%%%%%%%%%%%%%%%%%%%%%%%%%%%%%%%%%%%%%%%%%%%%%%%%%%%%%%
% \vspace{-10pt}
\subsection{\textbf{Trainable Quanvolutional Layers in QuNNs.}} \label{sec:trainable_QLayers} 
% \vspace{-3pt}
The first step is to make the quanvolutional layers trainable. The majority of works in the literature utilizes these layers for efficient feature extraction and pass them to classical layers, however, the quanvolution operation itself is not trainable.  
In order to enhance the adaptability of these layers within QuNNs it is important to make these layers trainable to harness their full potential.
Making the quanvolutional layer trainable entails making the quantum circuit (labeled as \circled{1} in Figure \ref{fig:methodology}) trainable.  
We do this by making the quanvolution layer as a pytorch layer which is later used as a standard layer during the model construction.  
The trainable QuNN utilized in this study is depicted in purple shaded region of Figure \ref{fig:methodology}, and operates through the following steps.

\begin{enumerate}%[leftmargin=-.06in]
    % \subsection*{\textbf{Encoding Classical Data into Quantum States}}
    \item \textbf{\textit{Encoding Classical Data into Quantum States.}} We use a kernel of size $2\times2$ which slides over the entire input image with stride 2. This results in $4$ classical features extracted after every step, i.e., \(x = \{x_1, x_2, x_3, x_4\}\). These features are then encoded into quantum states, specifically into the rotation angles of $RY$ gates. This approach is commonly called as angle encoding \cite{kashif:2021_DSE}.
    Each qubit $\ket{q_i}$ is prepared in the ground state $\ket{0}$, and the $RY$ gate applies a rotation by $x_i$ radians on the Bloch sphere, which can be described by the following Equation:

    \begin{equation} \label{eq:enc}
        U_{{\text{Enc}}_1} = |q_i\rangle = RY(x_i)|0\rangle \quad   \text{where} \quad  RY(\theta) = 
\begin{pmatrix}
\cos(x/2) & -\sin(x/2) \\
\sin(x/2) & \cos(x/2)
\end{pmatrix}
\end{equation}

    A total of $4$ qubits will be required, for encoding $4$ input features: 
    \begin{equation} \label{eq:enc1}
        U_{{\text{Enc}}_1} = \bigotimes_{i=1}^{4} RY(x_i) \ket{0}
    \end{equation}
    %
    %   
    % \vspace{-15pt}
   % $$ RY = e^{-i\phi\sigma_y/2} = \begin{bmatrix}\cos(\phi/2) & -\sin(\phi/2) \\
   %  \sin(\phi/2) & \cos(\phi/2) \\
   %  \end{bmatrix} $$
    
     \item \textbf{\textit{Trainable Quantum Circuit Operations.}}
    The encoded quantum states are then subjected to unitary evolution via an \(RX\) rotation gate. Let \(\theta_i\) be the angle of rotation for the \(i\)-th qubit applied by the \(RX\) gate:
    \begin{equation} \label{eq:U_rotation}
        U_{\text{Rotation}} = RX(\theta_i)|q_i\rangle  \quad \text{where} \quad RX(\theta) = 
\begin{pmatrix}
\cos(\theta/2) & -i\sin(\theta/2) \\
-i\sin(\theta/2) & \cos(\theta/2)
\end{pmatrix},
    \end{equation}
  For $4$-qubit layer, the above equation can written as: 
   
    \begin{equation}
        U_{\text{Rotation}} = \bigotimes_{i=1}^{4} RX(\theta_i) 
    \end{equation}
    To introduce entanglement between qubits in the quanvolutional layers, CNOT gates are applied between nearest neighbor qubits. Since we use total of $4$ qubits, the entanglement operation in this context can be defined as: 

    \begin{equation} \label{eq:CNOT}
        U_\text{{Entangle}} =  \bigotimes_{j=1}^3 CNOT_{j, j+1}
    \end{equation}

    % $$CNOT_{i,i+1}|q_iq_{i+1}\rangle = |q_i\rangle \otimes X^{q_i}|q_{i+1}\rangle$$
    
   \item  \textbf{\textit{Overall Transformation for Quantum Circuit.}}
    Combining the above steps, the overall transformation \(U_{\text{Circuit}_1}\) applied by the quanvolutional layer can be expressed as:

     \begin{equation}
        U_{\text{circuit}_1} = \left( \bigotimes_{i=1}^{3} CNOT_{i,i+1} \right) \cdot \left( \bigotimes_{i=1}^{4} RX(\theta_i) \right) \cdot \left( \bigotimes_{i=1}^{4} RY(x_i) \right)
       \end{equation}
       
    \begin{equation} \label{eq:final_Ucircuit}
        U_{\text{circuit}_1} =(U_{\text{Entangle}}.U_{\text{Rotation}}.U_{{\text{Enc}}_1})
    \end{equation}

In the proposed architecture, the first quanvolutional layer (QCL1) is designed with a circuit depth of $4$, while the second quanvolutional layer (QCL2) exhibits a reduced depth of $1$. Both QCL1 and QCL2 incorporate identical quantum operations, consisting of single-qubit parameterized rotation gates and two-qubit CNOT entangling gates, as illustrated in label \circled{1} of Figure~\ref{fig:methodology}. The use of only single and two-qubits gates makes the quantum circuits compatible with NISQ devices. The distinction between the two layers, i.e., QCL1 and QCL2 lies solely in the number of repetitions of the quantum circuit, thereby differentiating their respective depths.
This configuration (deep quantum layers in QCL1 and shallow layers in QCL2) is strategically chosen to distinctly evaluate the impact on performance when gradients from both layers are accessible in ResQuNNs, compared to the scenarios where the gradients of only the second layer are accessible. The reason is that when QCL1 is configured with a reduced depth, or when both QCL1 and QCL2 have equal depths, the effect of QCL1 on resulting performance difference may not be as pronounced or readily observable, even in cases where gradients from the second layer are accessible.
Applied to the initial state \(|0000\rangle\), the final state after the first QCL can be expressed as:

    \begin{equation} \label{eq:QCL1}
        \ket{\psi_{\text{QCL}_1}} = \prod_{l=1}^4U_{\text{circuit}_1}\ket{0000}
    \end{equation}

  where $l$ represents the total number of layers, i.e., number of times the circuit is repeated before measurement. The size of output after every quanvloutional layers is detailed in Table \ref{tab:output_size}. $\ket{\psi_{\text{QCL1}}}$ is the final state state after the first quanvolutionla layer before measurement and upon measurement the quantum state collapses and produce a classical result:
    \begin{equation}
        y = \mathcal{M}\ket{U_{\text{QCL1}}}
    \end{equation}
    where $\mathcal{M}$ denotes the qubit measurement in computational basis. Since the result from first quanvolutional layer is classical, we need to encode again before passing it to the second quanvolutional layer using the same equation (Eq. \ref{eq:enc}):
    
    \begin{equation}
        U_{{\text{Enc}}_2} = |q_i\rangle = RY(y_i)|0\rangle
    \end{equation}

    The kernel size remains the same so a total of 4 input features will be encoded from in every step:
    \begin{equation} \label{eq:enc2}
        y_\text{enc}=U_{{\text{Enc}}_2} = \bigotimes_{i=1}^{4} RY(y_i) 
    \end{equation}
    The rest of the operations of trainable quantum circuit in QCL2 are same as above and the final circuit equation will be:

    $$ U_{\text{circuit}_2} = \left( \bigotimes_{i=1}^{3} CNOT_{i,i+1} \right) \cdot \left( \bigotimes_{i=1}^{4} RX(\theta_i) \right) \cdot \left( \bigotimes_{i=1}^{4} RY(y_i) \right)$$
      
    \begin{equation} \label{eq:final_Ucircuit}
        U_{\text{circuit}_2} = U_{\text{Entangle}}.U_{\text{Rotation}}.U_{{\text{Enc}}_2}
    \end{equation}   

    The depth of quantum circuit in QCL is set to $1$. Hence, the final quantum state after the application of QCL2 will be:
    \begin{equation}
        \ket{\psi_{\text{QCL}_2}} = U_{\text{circuit}_2} \ket{y_\text{enc}}
    \end{equation}
    
 \begin{equation}
     Output_{\text{Quanv}} = \mathcal{M}\ket{U_{\text{QCL2}}}
 \end{equation}

   %  The overall of both quanvolutional layers can now be described as:
    
   % \begin{equation} \label{eq:final_unitary}
   %     |\psi_{\text{final}}\rangle = U_{\text{QCL2}} \cdot U_{\text{QCL1}}.U_{\text{E}_1}|0000\rangle 
   % \end{equation}
    
    % \item The encoded data then undergoes a unitary transformation, which is achieved through a PQC. We call this PQC, \emph{quanvolutional layer} throughout this paper, which is made trainable. The quanvolutional layer has $RX$ parameterized gate applied on every qubit, which can be described by the following Equation: 

    % $$ RX = e^{-i\phi\sigma_x/2} = \begin{bmatrix}\cos(\phi/2) & -i\sin(\phi/2) \\
    % -i\sin(\phi/2) & \cos(\phi/2) \\
    % \end{bmatrix} $$

    % Afterwards, the nearest neighbor qubits are entangled using \emph{CNOT} gate, which is represented by the following matrix:
    % \[ \text{CNOT} = \begin{pmatrix} 
    % 1 & 0 & 0 & 0 \\ 
    % 0 & 1 & 0 & 0 \\ 
    % 0 & 0 & 0 & 1 \\ 
    % 0 & 0 & 1 & 0 
    % \end{pmatrix} \]

    \item The measurement yields a list of classical expectation values. Similar to a classical convolution layer, each expectation value corresponds to a separate channel in a single output pixel.

    \item The outputs from quanvolutional layer are then postprocessed individually using both classical layer and quantum circuit.  
    
    \subsubsection{\textbf{Classical layer to postprocess quanvolutional layer's output}} First, we use classical layer to postprocess the output of quanvolutional layer, as highlighted in label \circled{2} in Figure \ref{fig:methodology}).  
    This  is a major practice in state-of-the-art QuNN architectures, where a fully connected neuron layer is used to take the output of the quanvolutional layer and produces the final classification output. 
    The number of neurons in this layer are equal to the total number of classes in the dataset used, which in our case is $10$ since we used MNIST handwritten digit dataset which has $10$ classes (more details in Section \ref{sec:exp_setup}).

    Let's assume the output of the quanvolutional layer is a set of measurements $\{ M =m_1, m_2 \ldots m_n \}$, where each $m_i$  represents the measurement of a qubit in terms of classical information, such as the expectation values in computational basis. This result is transformed to a vector format and then passed to the fully connected layer with $10$ neurons. 
    The output feature size from quanvolutional layers is different and depends on the kernel size and stride similar to that of classical convolutional layers. We experiment with a kernel size of $2\times 2$ and stride of $2$. The output feature size after each quanvolutional layer for different residual configuration (label \circled{5} in Figure \ref{fig:methodology}) is presented in the Table \ref{tab:output_size}.

    \begin{table}[htbp]
        \centering
        \caption{Output Feature Sizes for Two Quanvolutional Layer}
        % \hspace{15pt}
        \begin{tabular}{|p{5cm}|p{5cm}|p{5cm}|}
        \hline
        \hfil\textbf{Input Size}  & \hfil\textbf{Residual Configuration} & \hfil\textbf{Output Feature Size}  \\
        
            \hline
        \hfil$28\times 28\times  1$  & \hfil No residual &\hfil $7\times 7\times 1$ \\
        \hline
        \hfil$28\times 28\times  1$  &\hfil $X+O1$ &\hfil $14\times 14\times 1$ \\
        \hline
        \hfil$28\times 28\times  1$  &\hfil $O1+O2$ &\hfil $14\times 14\times 1$ \\
        \hline
        \hfil$28\times 28\times  1$  &\hfil $X+O2$ & \hfil $28\times 28\times 1$ \\
        \hline
        \hfil$28\times 28\times  1$  &\hfil $(X+O1)+O2$ &\hfil $28\times 28\times 1$ \\
        \hline

        \end{tabular}
        \label{tab:output_size}
    \end{table}
    
    The classical neural network layer with $10$ neurons is a fully connected layer, where each neuron receives inputs from all $n$ measurements of the quanvolutional layer and processes these inputs to produce an output. The output of $j$-th neuron in the classical layer can be described by the following Equation:

        $$y_j = f\left(\sum_{i=1}^{n} w_{ji} \cdot m_i + b_j\right)$$
where:
\begin{itemize}
  \item \(f\) is the SoftMax activation function, which converts a vector of K real numbers into a probability distribution of K possible outcomes. It is frequently used in multiclass classification problems.
  \item \(w_{ji}\) is the weight from the \(i\)-th input to the \(j\)-th neuron.
  \item \(b_j\) is the bias term for the \(j\)-th neuron.
  \item \(y_j\) is the output of the \(j\)-th neuron.
\end{itemize}

The operation of this layer can be compactly represented in matrix form as:

$$Y = f(W \cdot M + B)$$

where:
\begin{itemize}
  \item \(Y\) is the output vector \([y_1, y_2, ..., y_{10}]\).
  \item \(f\) is applied element-wise to the result.
  \item \(W\) is the weight matrix.
  \item \(M\) is the vector of measurements from the quantum layer.
  \item \(B\) is the bias vector.
\end{itemize}
    
\subsubsection{\textbf{Quantum Circuit to postprocess quanvolutional layer's output}}
We also use quantum circuit to postprocess the result from quanvolutional layer, as highlighted by label \circled{3} in Figure \ref{fig:methodology}. This is primarily for two reasons: firstly to demonstrate the working of fully quantum QuNNs, and secondly, to better understand the importance of residual connections by removing the influence of classical layers being used at the end.
The use of a quantum circuit for post-processing introduces certain computational overheads, mainly due to the prerequisite of re-encoding the (classical) outputs from the quanvolutional layers, a process that entails considerable computational expense. Also, the following trainable quantum circuit for postprocessing quanvolutional layer's results after the encoding, takes longer to produce the results than a simple classical neuron layer. We use $4$ classes from MNIST dataset for experimental simplification. 
For the purpose of encoding the outputs from the quanvolutional layer, \emph{Amplitude Encoding} \cite{LaRose:2020}, is used which can be described by the following Equation:
    \begin{equation} \label{eq1}
        \ket{\psi_x} = \sum_{i=1}^{2^n} x_i\ket{i}\\
    \end{equation}
    
where $x_i$ is the $i^{th}$ element of $x$ and $\ket{i}$ is the $i^{th}$ computational basis state. We used amplitude encoding for postprocessing the results of quanvolutional layers primarily because it allows to encode $2^n$ features in $n$ qubits \cite{kashif:2021_DSE}. 
Based on Table \ref{tab:output_size}, the smallest output feature size from quanvolutional layer(s) is $7\times 7\times 1 = 49$, which requires atleast $49$ qubits via angle encoding (one qubit per feature) and only $6$ qubits via amplitude encoding. Similarly, across the range of experiments we performed, the maximum output size of quanvolutional layer is $28\times 28\times 1 = 784$, requiring $784$ qubits, if encoded in qubit rotation angles, whereas amplitude encoding reduces the requirement to merely $10$ qubits. The use of $784$ qubits is not practical in current NISQ devices, therefore we use amplitude encoding to encode the qunavolutional layer's result for further processing by a quantum circuit.
To ensure a consistent impact of the quantum circuit's post-processing across all residual configurations, a $10$-qubit circuit is utilized for processing the outputs of the quanvolutional layer, corresponding to the maximum feature size of $784$. However, as mentioned earlier, we use $4$ classes, therefore, only $4$ qubits are measured each corresponding to one class. 
    The output of quanvolutional, once encoded, is subjected to the unitary evolution through a simple quantum circuit, which can be described by the following Equation: 

    \begin{equation}\label{eq:train_PQC}
    U(\theta) = \bigotimes_{j=1}^{9}\left(CNOT_{j,j+1}\right).\bigotimes_{i=1}^{10}R_y(\theta)
\end{equation}
    
    \end{enumerate}

% optimizing multiple quanvolutional layers presents a significant challenge that is One of the main difficulties lies in efficiently accessing gradients across these layers during training, which are essential in gradient-driven optimization techniques. We typically notice that when we train two or more quanvolutional layers, the gradients of only the last layer are accessible. We have shown these results as a primary motivation of this paper in Section \ref{sec:motiv_anaysis}.

% \vspace{-13pt}
\subsection{\textbf{Proposed Residual Quanvolutional Neural Networks (ResQuNNs).}}
We observe that optimizing multiple (two or more) quanvolutional layers presents a significant challenges. One of the main difficulties lies in gradients accessibility across these layers during training, which are essential in gradient-driven optimization techniques. We typically notice that when we train two or more quanvolutional layers, the gradients of only the last layer are accessible. We have shown these results as a primary motivation of this paper in Section \ref{sec:motiv_anaysis}.
To overcome the gradient accessibility issue encountered in multi-layered QuNNs, we introduce ResQuNNs. Drawing inspiration from classical residual neural networks, ResQuNNs incorporate residual blocks between quanvolutional layers. The residual blocks combine the output of existing quanvolutional layers with their own or the previous layer's input before forwarding it to the subsequent quanvolutional layer, as shown in label \circled{4} in Figure \ref{fig:methodology}. These residual connections play a pivotal role in facilitating comprehensive gradient access and, in turn, improving the overall training performance of QuNNs. 
The integration of residual blocks ensures that gradients flow efficiently across the network, thereby overcoming the challenges associated with training multiple quanvolutional layers.

% \vspace{-12pt}
\subsection{\textbf{Strategic Placement of Residual Blocks.}}
Within the proposed ResQuNN architecture, not every random residual configuration results in comprehensive gradient flow throughout the network, and the residual blocks needs to be strategically placed between quanvolutional layers. 
In our empirical investigation, we conduct extensive experimentation to determine the appropriate locations for inserting residual blocks within ResQuNNs comprising two quanvolutional layers, as shown in Figure \ref{fig:methodology}, labeled \circled{5}. 
This analysis helps us uncover the most effective residual configurations for ResQuNNs, enabling the gradients to flow throughout the proposed ResQuNN architecture, resulting in a deeper architecture and eventually the enhanced training performance. 
Based on the results of two quanvolutional layers, only the most effective configurations, i.e., those facilitating gradient flow through both layers were chosen for subsequent examination within three-layer quanvolutional frameworks, as denoted by label \circled{6} in Figure \ref{fig:methodology}.
We then trained ResQuNNs with two quanvolutional layers for all residual configurations for a multiclass classification problem (more on dataset in Section \ref{sec:exp_setup}) to assess the impact of gradient accessibility, both throughout the entire network (all quanvolutional layers) and within segments of the network (only the last quanvolutional layer), on the learning efficacy of QuNNs. 

%%%%%%%%%%%%%%%%%%%%%%%%%%%%%%%%%%%%%%%%%%%%%%%%%%%%%%%%%%%%%%%%%%%%%%%%%%%%%%%%%%%%%%%%%%%%%%%%%%%%%%%%%%%%%%%%%%%%%%%%%%%%%%%%%%%%%%%%%%%%%%%%%%%%%%%%%%%%%%%%%%%%%%%%%%%%%%%%%%%%%%%%%%%%%%%%%%%%%%%%%%%%%%%%%%%%%%%%%%%%%%%%%%%%%%%%%%%%%%%%%%%%%%%%%%%%%%%%%%
\begin{table*}[htbp]
\centering
\caption{\footnotesize Dataset and Hyperparemeters Specifications. BS = batch size, Opt = optimizer and LR = learning rate}
\resizebox{\textwidth}{!}{%
\begin{tabular}{|c|c|c|c|c|c|c|c|c|c|c|c|}
\hline
 \multicolumn{4}{|c|}{\textbf{Dataset Specs}} & \multirow{2}{*}{\textbf{BS}} & \multirow{2}{*}{\textbf{Opt}} & \multirow{2}{*}{\textbf{LR}} & \multicolumn{3}{c|}{\textbf{Quanvolutional layer specs}} & \multicolumn{2}{c|}{\textbf{Postprocessing}} \\ 
 \cline{1-4}
 \cline{8-12}
\textbf{Total classes} & \textbf{Samples/class} & \textbf{Train Data} & \textbf{Test Data} & & & & \textbf{Encoding} & \textbf{Kernel size} & \textbf{\# of Qubits} &\textbf{Classical} & \textbf{Quantum} \\
\hline
10 & 200 & 1600 & 400 & 16 & Adam & 0.01 & Angle (RY) & $2 \times 2$ &4 & Dense layer (10 neurons) &-\\
\hline
4 & 200 & 640 & 160 & 16 & Adam & 0.01 & Angle (RY) & $2 \times 2$ &4 & - & 10-Qubit circuit (4 measured)\\
\hline

\end{tabular}%
}

\label{table:params_specs}
\end{table*}
% \vspace{-10pt}
\section{Experimental Setup} \label{sec:exp_setup}
\begin{figure}[htbp]
    \centering
    % \hspace{-1pt}
    % height=1.2in, width=2.9in
    \includegraphics[scale=0.5]{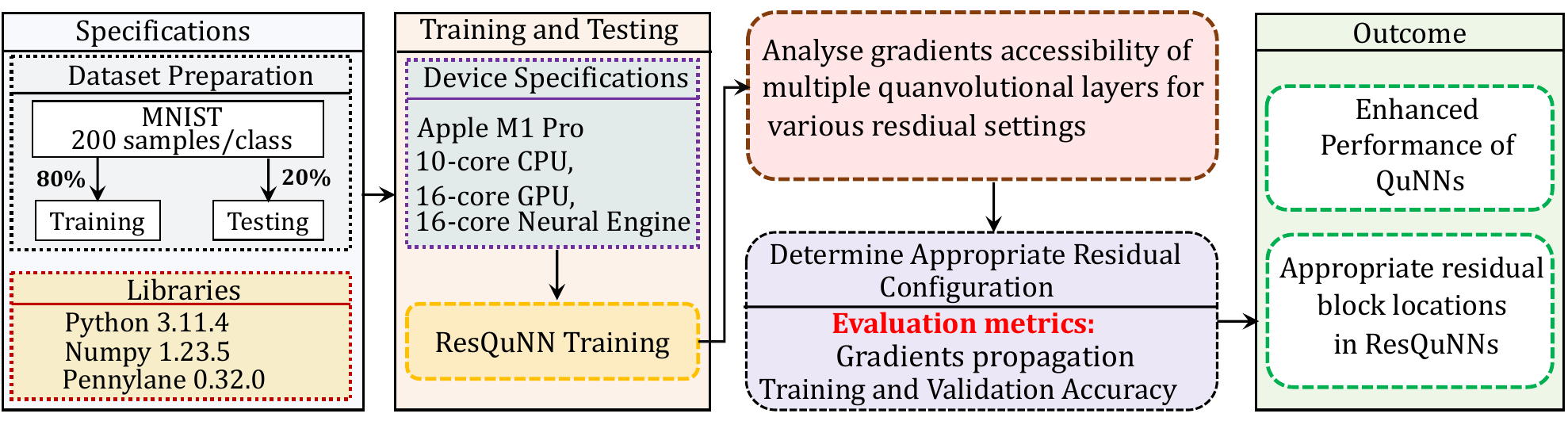}
    % \vspace{-5pt}
    \caption{\footnotesize Overview of Our Experimental Tool-flow}
    \label{fig:exp_workflow}
\end{figure}

An overview of the experimental toolflow utilized in this paper is depicted in Figure \ref{fig:exp_workflow}.
We trained the proposed ResQuNNs on a selected subset of the MNIST dataset. Table \ref{table:params_specs} summarizes the dataset and hyperparameters specification used in this paper.  The ResQuNNs are trained for $30$ epochs. Adam optimizer with learning rate of $0.01$ is used for the optimization.
The objective function chosen for training is cross entropy loss, which can be described (for C classes) by the following equation:

% \vspace{-10pt}

$$ H(p, q) = -\sum_{c=1}^{C} p(c) \log(q(c)) $$
% \vspace{-10pt}

where $H(p,q)$ is the cross entropy loss, $p(c)$ represents the true probability distribution for class $c$ and $q(c)$ denotes the predicted probability distribution for class.
All experiments are executed utilizing Pennylane, a Python library tailored for differentiable programming on quantum computing platforms \cite{Bergholm:2018}. 

% \vspace{-10pt}
%%%%%%%%%%%%%%%%%%%%%%%%%%%%%%%%%%%%%%%%%%%%%%%%%%%%%%%%%%%%%%%%%%%%%%%%%%%%%%%%%%%%%%%%%%%%%%%%%%%%%%%%%%%%%%%%%%%%%%%%%%%%%%%%%%%%%%%%%%%%%%%%%%%%%%%%%%%%%%%%%%%%%%%%%%%%%%%%%%%%%%%%%%%%%%%%%%%%%%%%%%%%%%%%%%%%%%%%%%%%%%%%%%%%%%%%%%%%%%%%%%%%%%%%%%%%%%%%%%
% \vspace{-18pt}

\section{Results and Discussion}
% In this section, we present the outcomes resulting from our experimental investigations, meticulously organized and subjected to thorough analysis, with the aim of deriving meaningful conclusions. 
% Our experimental scope encompassed the exploration of both two and three trainable quanvolutional layers. 

%Furthermore, we conducted a series of experiments to assess various configurations of residual blocks within the context of two and three quanvolutional layers to determine the optimal placement of these residual blocks.

%%%%%%%%%%%%%%%%%%%%%%%%%%%%%%%%%%%%%%%%%%%%%%%%%%%%%%%%%%%%%%%%%%%%%%%%%%%%%%%%%%%%%%%%%%%%%%%%%%%%%%%%%%%%%%%%%%%%%%%%%%%%%%%%%%%%%%%%%%%%%%%%%%%%%%%%%%%%%%%%%%%%%%%%%%%%%%%%%%%%%%%%%%%%%%%%%%%%

\subsection{\textbf{Two Quanvolutional Layers}}

We use kernel size of $2\times 2$, which results in  $4$-qubit quanvolutional layers. Our experimental framework involved extensive testing of all residual configurations as outlined in label \circled{3} of Figure \ref{fig:methodology}.
For each residual configuration, we investigated the gradient propagation through the quanvolutional layers. Afterwards, we trained the ResQuNNs for all these configurations to assess the impact of gradient accessibility throughout the network's layers.

%%%%%%%%%%%%%%%%%%%%%%%%%%%%%%%%%%%%%%%%%%%%%%%%%%%%%%%%%%%%%%%%%%%%%%%%%%%%%%%%%%

\subsubsection{\textbf{Analysis of Gradients Accessibility}}
{In the case of \emph{No Residual} setting (which refers to the configuration in which no residual blocks are incorporated within the multi-layered QuNNs and serves as a benchmark for comparison) we have already demonstrated in Figure \ref{fig:motiv_analysis_Grads} that the gradients are exclusively accessible for the second quanvolutional layer.
Therefore, we now proceed to investigate the gradient propagation through the quanvolutional layers for all other residual settings (from label \circled{5} in Figure \ref{fig:methodology}), the results of which are illustrated in Figure \ref{fig:2QCL_grad_results}. 
It is worth mentioning here that while the gradient accessibility results are presented with quantum layer postprocessing, the accessibility of gradients remains the same across layers with classical postprocessing as well.
\begin{figure}[htbp]
    \centering
    \includegraphics[scale=0.42]{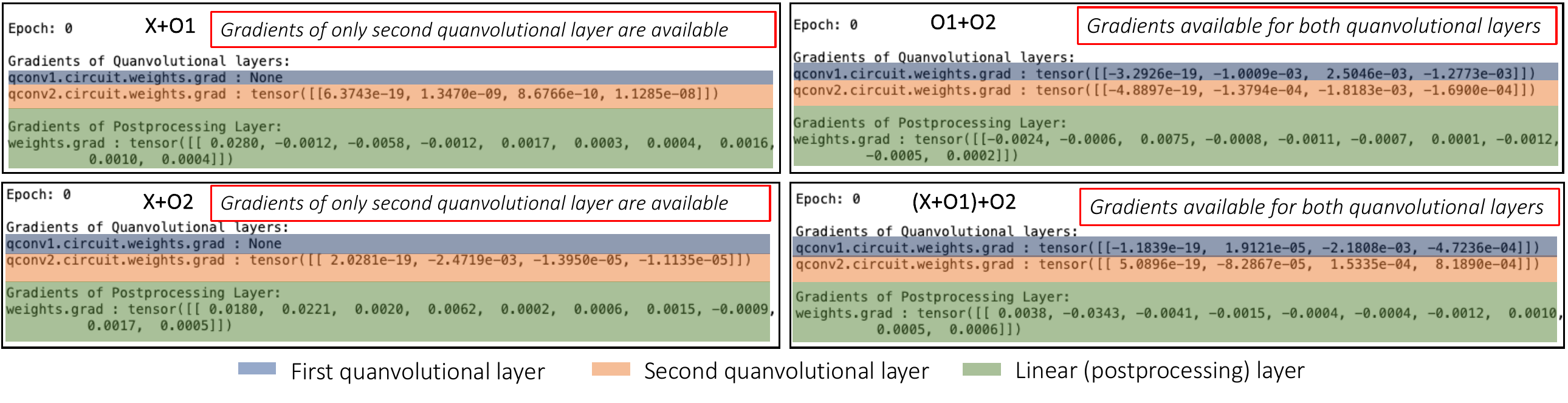}
    % \vspace{-5pt}
    \caption{\footnotesize Gradients Propagation Through the Network for Two Quanvolutional Layers for Different Residual Configurations. X=Input, O1, O2=Output of first and second quanvolutional layer respectively. %To demonstrate gradients accessibility, we consider each quanvolutional layer with a quantum circuit depth of $1$ since gradient propagation is not effectred by depth.
    }
    \label{fig:2QCL_grad_results}
\end{figure}
Our analysis reveals that specific residual configurations, notably $O1+O2$ and $(X+O1)+O2$, significantly enhance gradient propagation through the quanvolutional layers. This improvement not only makes them more conducive to gradient-based optimization strategies but also facilitates deeper learning within Quantum Neural Networks (QuNNs). In contrast, certain configurations, such as $X+O1$ and $X+O2$, mirror the conditions of a \emph{no residual} setting, where gradients fail to propagate past the terminal quanvolutional layer. Consequently, these configurations may exhibit a constrained training potential, impacting their effectiveness in QuNNs.
We observe that certain residual configurations, specifically $O1+O2$ and $(X+O1)+O2$, significantly enhance gradient propagation through the quanvolutional layers by facilitating the propagation of gradients through both the quanvolutional layers. This improvement not only makes them more conducive to gradient-based optimization strategies but also facilitates deeplearning within QuNNs.  
In contrast, other residual configurations, namely $X+O1$ and $X+O2$, are similar to the \emph{no residual} setting, wherein the gradients fail to propagate beyond the last quanvolutional layer in the network. 
Consequently, these configurations may exhibit a constrained training potential.

%%%%%%%%%%%%%%%%%%%%%%%%%%%%%%%%%%%%%%%%%%%%%%%%%%%%%%%%%%%%%%%%%%%%%%%%%%%%%%%%%%%%%%%

\subsubsection{\textbf{Analysis of Training Performance}}

We then trained the proposed ResQuNNs on a subset of MNIST dataset, as detailed in Section \ref{sec:exp_setup}, for all the residual configurations labeled as \circled{3} in Figure \ref{fig:methodology}, separately with classical and quantum postprocessing. 
The experimental procedures are consistent across all trials, adhering to the parameters discussed in Section \ref{sec:exp_setup}. It is worth mentioning here that the \emph{No residual} configuration also serves as a common benchmark for all the other residual configurations to analyze whether or not the inclusion of residual connections bring any advantage. 
Below, we separately discuss the training results of ResQuNNs both with classical and quantum postprocessing. 

%%%%%%%%%%%%%%%%%%%%%%%%%
\begin{enumerate}
    
\item \textbf{Classical Postprocessing.}
We first evaluate the training performance of ResQuNNs that utilize classical neuron layers to postprocess the output of quanvolutional layers, as highlighted by label \circled{2} in Figure \ref{fig:methodology}. The training results of ResQuNNs with classical postprocessing are shown in Figure \ref{fig:res_2by2_2QCL}.
A common and a key observation is that the incorporation of residual connections consistently improves the training performance of QuNNs, regardless of the specific residual configuration. 
For instance, while only the gradients of second quanvolutional layer are accessible in both \emph{No residual} and $X+O1$ configurations, the introduction of a residual connection in $X+O1$ leads to superior performance compared to the \emph{No residual} approach.
We now individually explore the impact of different residual configurations on the performance of QuNNs with a focus on training efficiency and generalization capabilities (validation accuracy).
% in case of two quanvolutional layers employing $2\times2$ kernel. 
% As discussed earlier, our examination in case of two quanvolutional layers spans several configurations, categorized and analyzed as follows:

% This is demonstrated in Figure \ref{fig:2QCL_grad_results} (label \circled{1} to \circled{4}), where configurations like $X+O1$ perform better that no residual configuration, even though both provide access only to the second quanvolutional layer's gradients.
%
%
\begin{figure*}[htbp]
    % \centering
    % \hspace{-8pt}
    % height=0.8in, width=5.8in
    \includegraphics[scale=0.5]{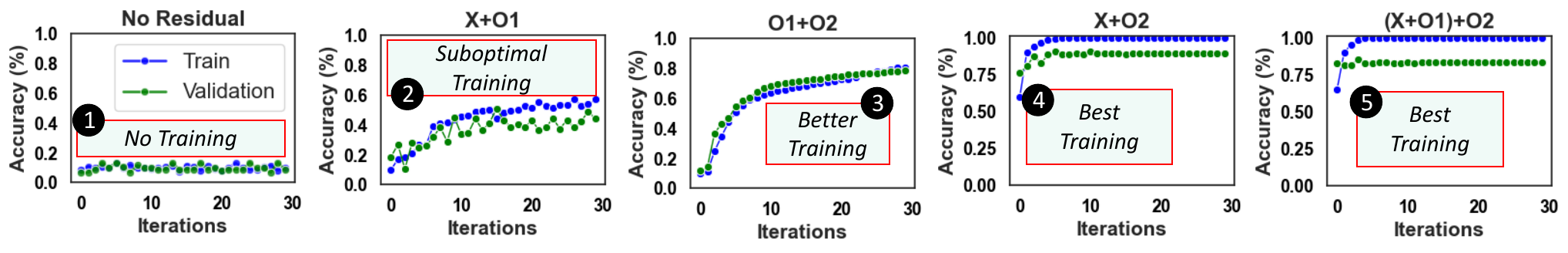}
    % \vspace{-20pt}
    \caption{\footnotesize {Two Quanvolutional Layers - Training results of all residual configurations with classical postprocessing and $2\times 2$ Kernel. No Residual setting refers to multi-layered QuNNs without any residual block} }
    \label{fig:res_2by2_2QCL}
\end{figure*}

\begin{itemize}
    \item \textbf{\textit{No Residual.}}
    When there is \emph{No} residual block, the QuNNs exhibit \emph{no training} as shown in label \circled{1} of Figure \ref{fig:res_2by2_2QCL}. Although, even in case of no residual, the gradients of second quanvolution layer are available but a significant information is lost in the first quanvolution layer (firstly because of the $2\times 2$ quanvolution filter, and secondly due to the measurements and re-encodings), which does not contribute to the learning because of no gradients.   

    \item \textbf{\textit{X+O1.}} When the residual block is added after the first qunavolutional layer, i.e., the input is added to the output of first quanvolutional and then passed as input to the second quanvolutional layer, the model shows significant performance enhancement with improved training and validation accuracy compared to \emph{No residual} setting, as shown in label \circled{2} of Figure \ref{fig:res_2by2_2QCL}. 
    While $X+O1$ residual configuration marks a significant improvement over the \emph{No residual} configuration, it nonetheless presents a sub-optimal training performance as the first quanvolutional layer does not take part in the overall optimization because of no gradients. 

    \item \textbf{\textit{O1+O2.}} Contrary to the \emph{No residual} and $X+O1$ configurations, the residual setup of $O1+O2$ facilitates the propagation of gradients through both quanvolutional layers, which is considered more conducive for gradient-based optimization techniques. 
    This facilitation of gradient propagation results in a marked improvement in performance with the $O1+O2$ configuration, as depicted in label \circled{3} of Figure \ref{fig:res_2by2_2QCL}
    The training and generalization capabilities of the $O1+O2$ configuration significantly exceed those of $X+O1$, demonstrating approximately a \textbf{31\%} increase in training accuracy and a \textbf{45\%} enhancement in validation accuracy, respectively. This superiority can be attributed to the $O1+O2$ configuration's facilitation of gradient propagation through both quanvolutional layers, in contrast to $X+O1$, where gradient access is limited exclusively to the final quanvolutional layer.

    \item \textbf{\textit{X+O2 and (X+O1)+O2.}} The residual configurations where the input is propagated  directly ($X+O2$) or indirectly ($(X+O1)+O2$) to the output of second quanvolutional layer, the ResQuNNs outperform other residual configurations, as shown in \label{4} and \label{5} of Figure \ref{fig:res_2by2_2QCL}. Additionally, both these residual settings exhibits very similar performance.
    
\end{itemize}

We argue that the residual configurations facilitating gradient flow through both quanvolutional layers, specifically, $O1+O2$ and $(X+O1)+O2$, will inherently exhibit superior performance in terms of both training and generalization compared to those configurations where gradient access is confined to the last quanvolutional layer only (\emph{No Residual}, $X+O1$, and $X+O2$ configurations). This is evident form the performance comparison of $X+O1$ and $O1+O2$ (label \circled{2} and \circled{3} in Figure \ref{fig:res_2by2_2QCL}), where $O1+O2$ performs significantly better.
Additionally, we hypothesize that the superior performance of $X+O2$ and $(X+O1)+O2$ can be attributed to the last classical layer (used to meaningfully post process quanvolutional layer's output) in the network since the (input) classical data in both these settings is directly ($X+O2$) or indirectly ($(X+O1)+O2$) passed to the classical layer which tends to overshadow the learning contribution from quanvolutional layers. 
Therefore, in next section, we scrutinize the influence of classical layer (used for postprocessing) on the overall learning performance in all the residual configurations from Figure \ref{fig:methodology} label \circled{5}. This analysis is to identify the residual configurations yielding better training and generalization performance primarily due to the learning contributions from the quanvolutional layers.
 
%%%%%%%%%%%%%%%%%%%%%%%%%%%%%%%%%%%%%%%%%%%%%%%%%%%%%%%%%%%%%%%%%%%%%%%%%%%%%%%%%%%%%%%%%%%%%%%%%%%%%%%%%%%%%%%%%%%%%%%%%%%%%%%%%%%%%%%%%%%%%%%%%%%%%%%%%%%%%
\paragraph{\textbf{Comparison with Benchmark Models.}}
To assess the impact of the classical layer in various residual configurations of ResQuNNs, we conducted a detailed analysis using benchmark models. The benchmark models replicate the ResQuNNs with the residual configurations as shown in label \circled{5} of Figure \ref{fig:methodology}, with one key difference: the quanvolutional layers were made \emph{untrainable}, i.e., their weights are not updated during training, and only the classical layers were subjected to training. 
This exercise allows us to determine the extent to which the classical layer tends to dominate the learning process, compared to the quanvolutional layers across all the residual configurations. 
To put it simply, configurations yielding similar performance when the quanvolutional layer remains untrained and only the classical layer is trained owe their success primarily to the classical layer. Conversely, configurations that demonstrate improved performance when both the quanvolutional and classical layers are trained, as opposed to when only the classical layers are trained, benefit from the residual configuration and enhanced gradient flow throughout the network.

%benchmark comparison results figure copied here for appropriate place in PDF
\begin{figure*}[htbp]
    % \centering
    % \hspace{-0.99cm}
    % height=1.4in, width=5in
    \includegraphics[scale=0.57]{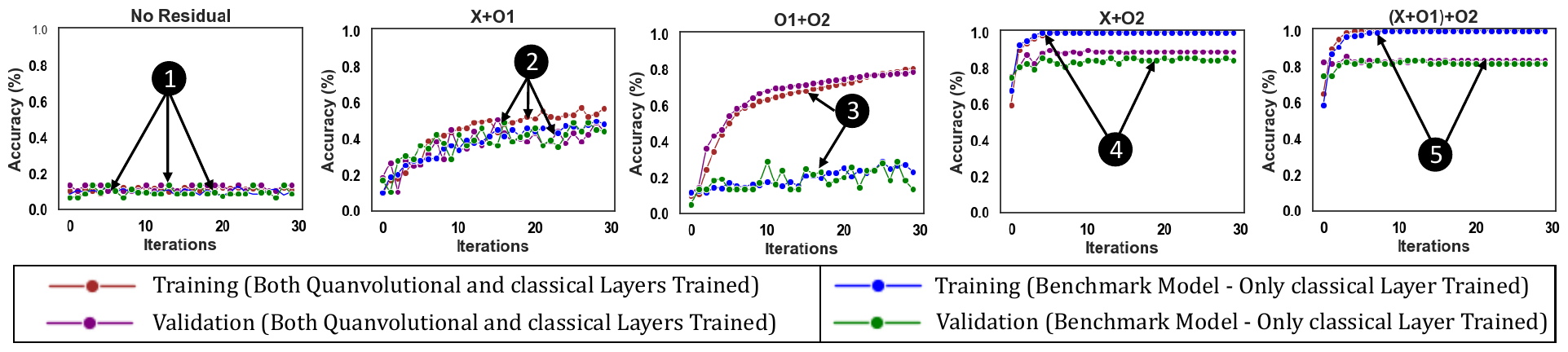}
    % \vspace{-20pt}
    \caption{\footnotesize {Two Quanvolutional Layers - Comparison of ResQuNNs with benchmark models for $2\times 2$ Kernel. No Residual setting refers to multi-layered QuNNs without any residual block}}
    \label{fig:res_2by2_2QCL_benchmark}
\end{figure*}

The benchmark models are trained using the same hyperparameters and dataset as detailed in Section \ref{sec:exp_setup}, and the results are compared with the models, where both quanvolutional and classical layers were trainable. 
The comparison results are presented in Figure \ref{fig:res_2by2_2QCL_benchmark}. The observed performance trends in these models gives insights into the effectiveness of quanvolutional layers in different residual configurations. \textit{The key findings from this comparative analysis are summarized below:}

%%%%%%%%%%%%%%%%%%%%%%%%%%%%%%%%%%%%%%%%%%%%%%%%%%%%%%%%%%%%%%%%%%%%%%%%%%%%%%%%%%%%%%%%
\begin{itemize}
  
    \item \textbf{\textit{No Residual.}}
    In the scenario where no residual connections are utilized, gradient flow is restricted exclusively to the second quanvolutional layer. Our observations indicate that the performance of the model, when both quanvolutional and classical layers undergo training, is very similar to that of the benchmark models in which training is restricted solely to the classical layers, with quanvolutional layers remaining static, as shown in label \circled{1} of Figure \ref{fig:res_2by2_2QCL_benchmark}. Nevertheless, in both settings, this particular configuration results in \emph{minimal} or \emph{No} training. 

%%%%%%%%%%%%%%%%%%%%%%%%%%%%%%%%%%%%%%%%%%%%%%%%%%%%%%%%%%%%%%%%%%%%%%%%%%%%%%%%%%%%%%%%

      \item\textbf{\textit{Residual Configuration $\textbf{X}\textbf{+}\textbf{O1}$.}}
    In this residual configuration, despite gradient accessibility being limited to the second quanvolutional layer, the introduction of residual connections significantly enhances the model's training capability. However, the comparative analysis of actual ResQuNNs (from Figure \ref{fig:res_2by2_2QCL}) with benchmark models suggests that, the performance in this very residual configuration is predominantly influenced by the classical layer at the end of the quanvolutional layers, as highlighted by label \circled{2} in Figure \ref{fig:res_2by2_2QCL_benchmark}.
This observation suggests that, within this configuration, the classical layer plays a critical role in the model's overall learning and performance outcomes.

%%%%%%%%%%%%%%%%%%%%%%%%%%%%%%%%%%%%%%%%%%%%%%%%%%%%%%%%%%%%%%%%%%%%%%%%%%%%%%%%%%%%%%%%%

      \item \textbf{\textit{Residual Configuration (O1+O2).}}
    This residual configuration potentially enables the deep learning by permitting gradient propagation through the quanvolutional layers, thus ensuring their active involvement in the learning process.
    This results in a significant enhancement in the model's performance, particularly when compared with the benchmark model. 
    As denoted by label \circled{3} in Figure \ref{fig:res_2by2_2QCL_benchmark}, the performance of the model when both the quanvolutional layers along with classical layer are trained surpasses that of the benchmark model (where only classical layer is trained and quanvolutional layers are kept frozen) by approximately \textbf{66\%} in training accuracy and \textbf{64\%} in validation accuracy. 
    This observation implies that in $O1+O2$ residual configuration, the learning performance is primarily driven by the quanvolutional layers. 
    Such a notable enhancement underscores the benefit of fully utilizing both quanvolutional layers when gradients are accessible for them, highlighting the importance of their contribution to the learning process.

%%%%%%%%%%%%%%%%%%%%%%%%%%%%%%%%%%%%%%%%%%%%%%%%%%%%%%%%%%%%%%%%%%%%%%%%%%%%%%%%%%%%%%%%%

      \item \textbf{\textit{Residual Configuration with Direct/Indirect Input to Classical Layer (X+O2 and (X+O1)+O2).}} 
    In scenarios where the input is either directly or indirectly routed to the classical layer at the network's end, such as in $X+O2$ and $(X+O1)+O2$, the performance of QuNNs with trainable quanvolutional demonstrates a close approximation to that of the benchmark models. This observation is highlighted by labels \circled{4} and \circled{5} in Figure \ref{fig:res_2by2_2QCL_benchmark}, indicating the predominant role of the classical layer in the learning process.
    The similar performance is achieved when only the classical layer is subjected to training, in comparison to scenarios where both quanvolutional and classical layers are trained, highlighting the significant role of the classical layer in the learning process in these residual configurations.
    However, considering the enhanced performance in the $O1+O2$ configuration, which allows gradient flow through both the quanvolutional layers, it suggests that $(X+O1)+O2$ could be effective in scenarios without a classical layer at the end. 
    This is because it also allows gradient propagation across all quanvolutional layers, potentially enabling the deep learning in QuNNs and eventually enhancing the learning performance, when the classical layer's dominance is removed. 
    To further elaborate on this, in the following section, we use the quantum circuit to postprocess the results of quanvolutional layers to further understand the role of residual approach in enabling deep learning in QuNNs, when there are no classical layers involved. 
\end{itemize}
%%%%%%%%%%%%%%%%%%%%%%%%%%%%%%%%%%%%%%%%%%%%%%%%%%%%%%%%%%%%%%%%%%%%%%%%%%%%%%%%%%%%%%%%%%%%%%%%%%%%%%%%%%%%%%%%%%%%%%%%%%%%%%%%%%%%%%%%%%%%%%%%%%%%%

%%%%%%%%%%%%%%%%%%%%%%%%%%%%%%%%%%%%%%%%%%%%%%%%%%%%%%%%%%%%%%%%%%%%%%%%%%%%%%%%%%%%%%%%%%
\item \textbf{Quantum Postprocessing}
In this section, we present the training results of ResQuNNs using quantum layer/circuit to postprocess the results of quanvolutional layer(s), as shown in label \circled{3} of Figure \ref{fig:methodology}. The primary reason behined quantum postprocessing is to reduce the influence of classical layer on the overall learning performance.
While using quantum postprocessing, we observe clear advantage of residual configurations that allow the gradient flow across both the layers of the network compared to those which allow partial gradient flow. The training results are depicted in Figure \ref{fig:res_2by2_2QCL_Q_End}. 
Below, we separately discuss the results of ResQuNN with different residual configurations (from label \circled{5} in Figure \ref{fig:methodology}), employing two quanvolutional layers.

% training results in Quantum postprocessing
\begin{figure*}[htbp]
    % \centering
    \hspace{-20pt}
    \includegraphics[scale=0.5]{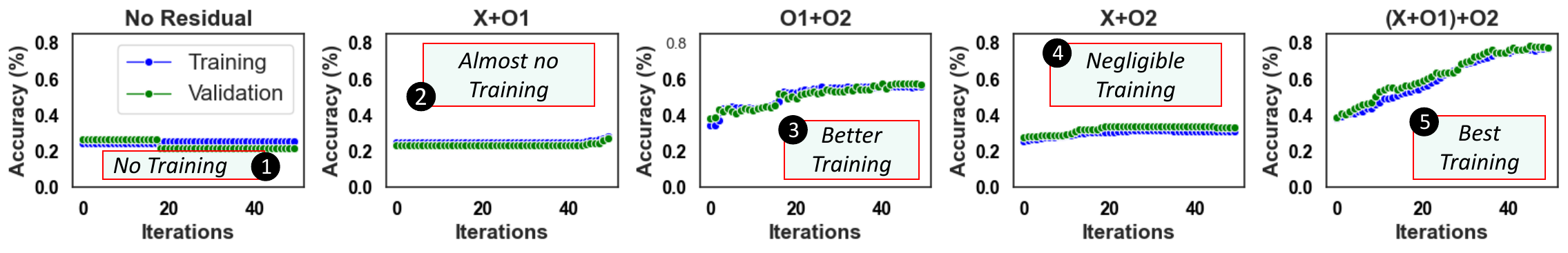}
    % \vspace{-20pt}
    \caption{\footnotesize {Two Quanvolutional Layers - Training results of all residual configurations with quantum postprocessing $\times 2$ Kernel. No Residual setting refers to multi-layered QuNNs without any residual block}}
    \label{fig:res_2by2_2QCL_Q_End}
\end{figure*}

\begin{itemize}
    \item \textbf{\textit{Configurations with Restricted Gradient Flow (No residual, X+O1, X+O2).}} The residual configurations that allow partial gradient flow, i.e, \emph{No residual, X+O1 and X+O2} exhibits no to negligible training, as shown in labels \circled{1}, \circled{2} and \circled{4} of Figure \ref{fig:res_2by2_2QCL_Q_End}. This outcome underscores the critical role of gradient flow in the training process of QuNNs.

    \item \textbf{\textit{Configurations Facilitating Full Gradient Flow (X+O2, (X+O1)+O2).}} The residual configurations that facilitate gradients flow through all the layers of the network specifically `$X+O2$' and `$(X+O1)+O2$', demonstrated significantly enhanced performance and completely outperforms other residual configurations, as illustrated by label \circled{3} and \circled{5} in Figure \ref{fig:res_2by2_2QCL_Q_End}.
\end{itemize}

The contrast in training performance across different residual configurations elucidates the tremendous importance of gradient flow in the architecture of ResQuNNs. 
Configurations enabling full gradient availability across network layers emerged as distinctly advantageous, siginificantly improving the training potential of ResQuNNs. 
Conversely, configurations with limited gradient flow manifested negligible training outcomes, highlighting the inadequacy of partial gradient availability for effective network learning. 
This distinction underlines the significance of architectural choices in the design of ResQuNNs, particularly in the context of integrating quantum and classical layers for enhanced learning capabilities.

% These findings contribute to the burgeoning field of quantum deep learning, offering insights into the architectural prerequisites for optimizing the training efficacy of quantum neural networks. Through the strategic incorporation of residual configurations facilitating comprehensive gradient flow, this study underscores the feasibility and advantages of deploying ResQuNNs for advanced quantum computing applications.
\end{enumerate}
%%%%%%%%%%%%%%%%%%%%%%%%%%%%%%%%%%%%%%%%%%%%%%%%%%%%%%%%%%%%%%%%%%%%%%%%%%%%%%%%%%%%%%%%%%%%%%%%%%%%%%%%%%%%%%%%%%%%%%%%%%%%%%%%%%%%%%%%%%%%%%%%%%%%%%%%%%%%%%%%%%%%%%%%%%%%%%%%%%%%%%%%%%%%%%%%%%%%%%%%%%%%%%%%%%%%%%%%%%%%%%%%

\subsection{Three Quanvolutional Layers}
In previous sections, we established the critical importance of gradient accessibility across all the layers in QuNNs for achieving better training performance. In this section, we extend our exploration to demonstrate that our proposed ResQuNNs can further facilitate the utilization of an increased number of quanvolutional layers for more deeper networks. 
Specifically, we consider the scenario of training QuNNs with three quanvolutional layers. 
% In a prior discussion (Figure \ref{fig:motiv_analysis_Grads}), we illustrated the case of \emph{No Residual} in three quanvolutional layers, revealing that gradients were only available for the final quanvolutional layer.
%
% Here, we examine the gradient results for other residual configurations. 
Due to space constraints, we only present the results for residual configurations where gradients are accessible across all three quanvolutional layers, as shown in Figure \ref{fig:3QCL_grads_residual}. 
Out of $15$ potential residual configurations examined, only two, specifically $(O1+O2)+O3$ and $((X+O1)+O2)+O3$, allowed gradient access across all three layers. The other configurations restricted gradient flow to one or two layers.
This further underscores the significance of our proposed residual approach, as it enables the desired comprehensive gradient accessibility, with greater number of layers, effectively enabling the deep learning in QuNNs, which can be pivotal for optimizing complex problems.
% \vspace{-10pt}

\begin{figure}[htbp]
    \centering
    % \hspace{-14pt}
    % height=2.2in, width=3.4in
    \includegraphics[scale=0.41]{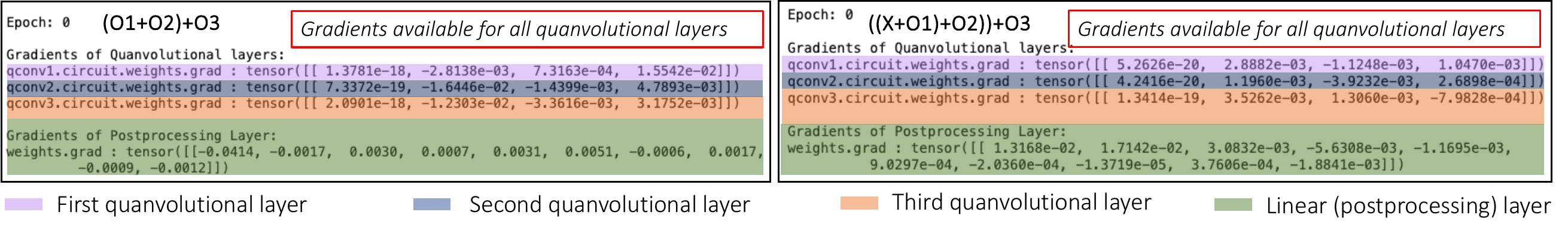}
    % \vspace{-7pt}
     \caption{\footnotesize Gradients Propagation Through the Network for Three Quanvolutional Layers for (O1+O2)+O3 (left) and ((X+O1)+O2)+O3 (right)}
    \label{fig:3QCL_grads_residual}
\end{figure}

%%%%%%%%%%%%%%%%%%%%%%%%%%%%%%%%%%%%%%%%%%%%%%%%%%%%%%%%%%%%%%%%%%%%%%%%%%%%%

\section{Conclusion}
In this paper, we presented a framework for developing Quanvolutional Neural Networks (QuNNs) by introducing trainable quanvolutional layers and the concept of Residual Quanvolutional Neural Networks (ResQuNNs) for enabling the deep learning in QuNNs. 
Our approach addresses the significant challenges of gradient flow and adaptability in deep QuNN architectures, enabling enhanced training performance and scalability.
A key contribution of this work is the development of trainable quanvolutional layers, which enhance the learning potential of QuNNs, and the strategic incorporation of residual blocks to ensure comprehensive gradient accessibility across the network in multi-layered QuNNs, thus, enabling deeper, more complex QuNNs architectures to be effectively trained.
Our empirical findings highlight the importance of optimal placement of residual blocks, providing critical insights for future QuNN architecture designs with even more deeper layers. 
%This advancement not only boosts the performance of QuNNs but also lays the groundwork for future quantum deep learning developments, potentially accelerating the application of quantum computing across various domains.
%
%
In summary, our work signifies a substantial leap in quantum deep learning, offering new directions for theoretical and practical quantum computing research. By tackling core issues of adaptability and gradient flow, our work paves the way for the development of more robust and efficient quantum neural networks, marking a pivotal step toward harnessing the full potential of quantum technologies.

% \begin{spacing}{0.85}

% \begin{spacing}[0.95]
    
% \section*{Acknowledgements}
% \vspace{-0.14cm}
% This work was supported in part by the NYUAD Center for Quantum and
% Topological Systems (CQTS), funded by Tamkeen under the NYUAD Research
% Institute grant CG008.

\section*{Data availability}
The datasets used and/or analysed during the current study available from the corresponding author on reasonable request.

% \section*{References}
% \bibliographystyle{}
\bibliography{main.bib}

\section*{Acknowledgements}
% \vspace{-0.14cm}
This work was supported in part by the NYUAD Center for Quantum and Topological Systems (CQTS), funded by Tamkeen under the NYUAD Research Institute grant CG008.

\section*{Author contributions}
M.K. and M.S. conceptualized and refined the research idea. M.K. implemented the idea, conducted the experiments, and generated the results, with M.S. providing supervision throughout the experimentation and implementation process. M.K. prepared the initial draft of the manuscript, while M.S. critically reviewed and edited the final version.

\section*{Competing Interests}
% \vspace{-0.14cm}
The authors declare no competing interests.

\end{document}